\documentclass[letterpaper, 10 pt, conference]{ieeeconf}  
\IEEEoverridecommandlockouts                              
\usepackage{graphics}    
\usepackage{times}       
\usepackage{amsmath}     
\usepackage{amssymb}     
\usepackage{graphicx}
\usepackage{algorithm}
\usepackage[noend]{algpseudocode}
\usepackage[font=footnotesize]{subfig}
\usepackage{color}
\usepackage{enumerate}
\usepackage{bm}
\usepackage{booktabs}

\usepackage[font=small]{caption}

\def\secref#1{Sec.~\ref{#1}}
\def\figref#1{Fig.~\ref{#1}}
\def\tabref#1{Tab.~\ref{#1}}
\def\eqref#1{Eq.~(\ref{#1})}

\newcommand\etal{\emph{et al.}}

\newcommand{\prob}{p}
\newcommand{\probcond}[2]{\prob(#1\,\mid\,#2)}

\newcommand{\var}[1]{\mathnormal{\MakeUppercase{#1}}}
\newcommand{\varset}[1]{\bm{\MakeUppercase{#1}}}

\newcommand{\graph}{\mathcal{G}}

\newcommand{\edge}{\var{e}}
\newcommand{\Edges}{\varset{E}}

\newcommand{\Vertex}{\var{v}}
\newcommand{\Vertices }{\varset{V}}

\newcommand{\ObservedEdges}{\varset{Z}}
\newcommand{\UnobservedEdges}{\varset{U}}

\newcommand{\mutualinfo}{\mathcal{I}}
\newcommand{\entropy}{\mathcal{H}}

\def\argmax{\mathop{\rm argmax}}

\title{\LARGE \bf Long-Term Robot Navigation in Indoor Environments\\ 
Estimating Patterns in Traversability Changes}

\author{Lorenzo Nardi \and Cyrill Stachniss
  \thanks{All authors are with the University of Bonn, Germany. }%
  \thanks{This work has partly been supported by the German Research Foundation under Germany's Excellence Strategy, EXC-2070 - 390732324 (PhenoRob).
  }%
}

\begin{document}
\maketitle
\thispagestyle{empty}
\pagestyle{empty}

\begin{abstract}

Nowadays, mobile robots are deployed in many indoor environments, such as offices or hospitals. These environments are subject to changes in the traversability that often happen by following repeating patterns. In this paper, we investigate the problem of navigating in such environments over extended periods of time by capturing these patterns and exploiting this knowledge to make informed decisions. Our approach incrementally estimates a model of the traversability changes from robot's observations and uses a probabilistic graphical model to make predictions at currently unobserved locations. In the belief space defined by the predictions, we plan paths that trade off the risk to encounter obstacles and the information gain of visiting unknown locations. We implemented our approach and tested it in different indoor environments. The experiments suggest that in the long run, our approach leads to navigation along shorter paths compared to following a greedy shortest path policy.

\end{abstract}

\section{Introduction}
\label{sec:intro}

Over the last decade, many mobile robots have been deployed in indoor environments such as offices, hospitals, and shopping malls. Most robot navigation systems rely on a static representation of the environment such as occupancy grid maps or topological maps for planning and navigating to targeted locations. In reality, mobile robots are often employed in environments that are subject to changes in traversability. Traditional navigation systems avoid obstacles by performing reactive strategies~\cite{fox1997jra} or planning local deviations~\cite{stachniss2002iros}. These approaches are effective to tackle unforeseen obstacles but have no memory about previously experienced situations. Thus, when encountering the same situation multiple times, the robot may perform every time the same sub-optimal behavior.

\begin{figure}[t]
  \vspace{-0.3em}
  \centering
  \subfloat[Working state.]{
    \includegraphics[width=0.415\linewidth]{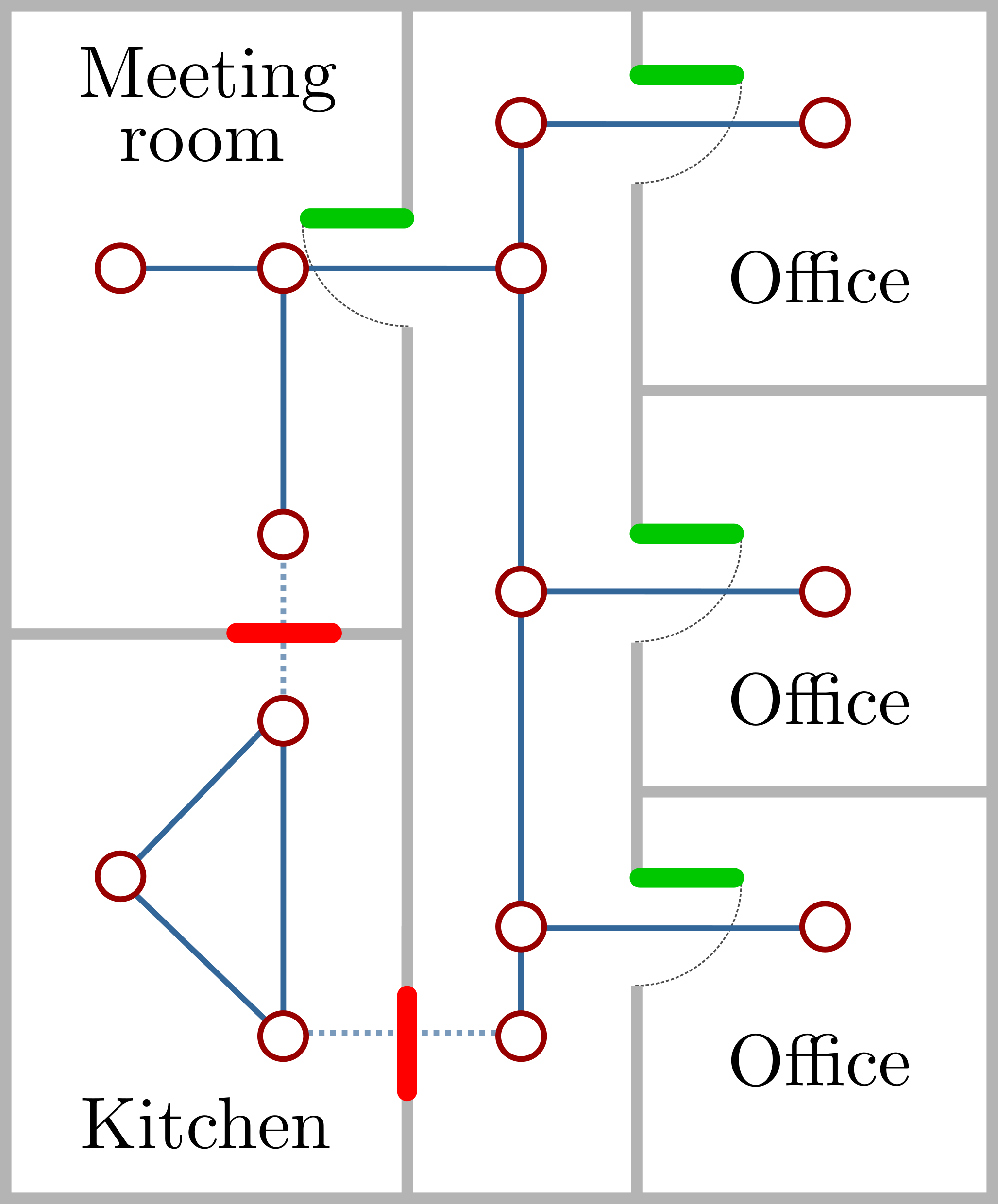}\label{fig:motivation:work}\hspace{9px}
  }
  \subfloat[Coffee break.] {
    \includegraphics[width=0.415\linewidth]{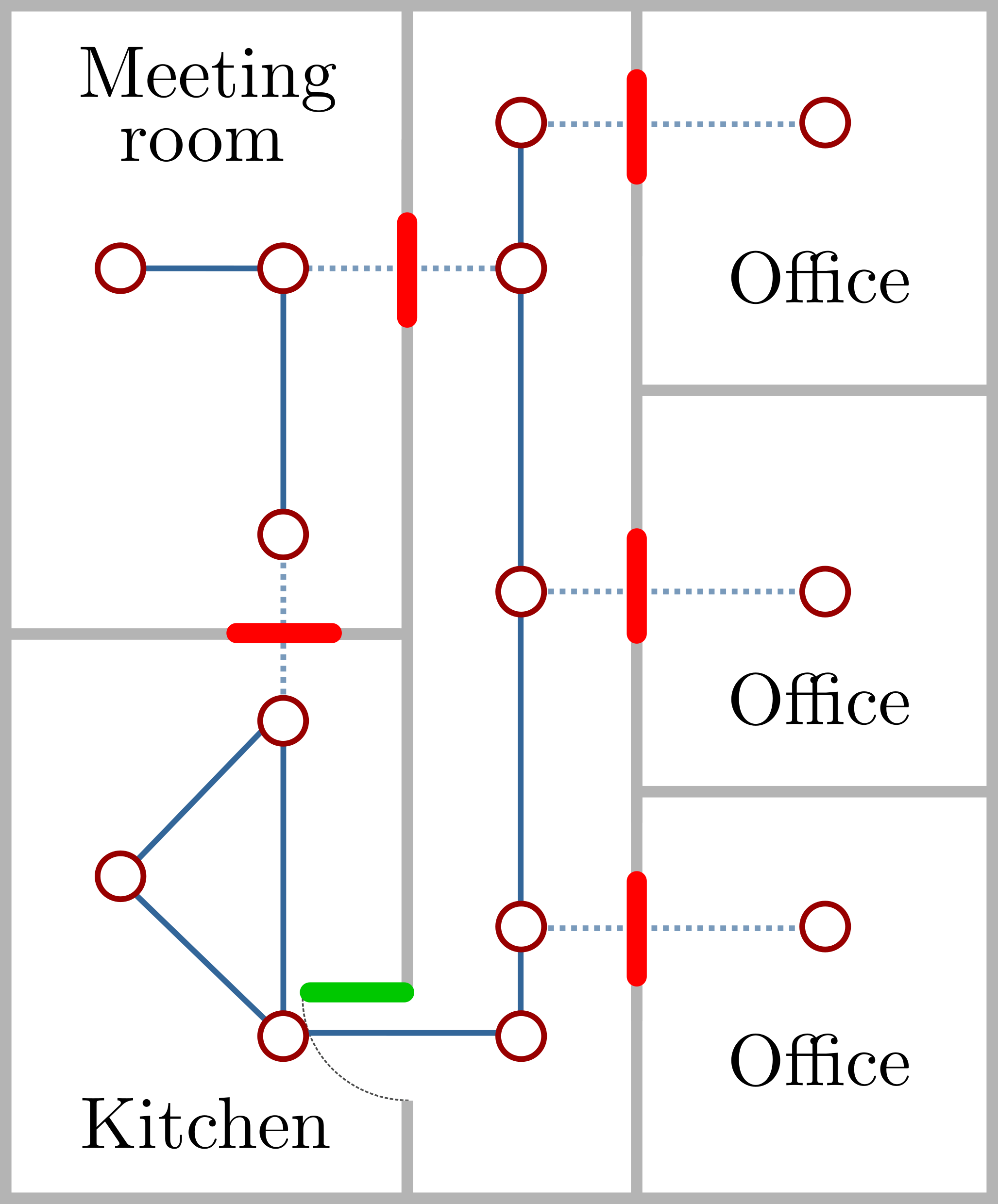}\label{fig:motivation:coffee}
  }
 \caption{Traversability change patterns on the topological map of an office, where the red circles are the nodes and the blue solid lines are the traversable connections among them.}
 \vspace{-0.5em}
\label{fig:motivation}
\end{figure}

In indoor environments, there are many changes in the traversability that happen following repeating patterns.
For example, the doors in an environment could be open or closed at the same time according to certain patterns.
In the environment illustrated in~\figref{fig:motivation:work}, the offices' doors are typically open~(green) while people are working.
Whereas, if the kitchen is open, it is likely that people are enjoying a coffee and so that the offices' doors are closed~(red), see~\figref{fig:motivation:coffee}.
When deploying a robot in such environments over a longer period of time, it can observe these repeating patterns and exploit this knowledge to navigate along shorter paths thus increasing the efficiency of its operations.

In this paper, we investigate the problem of (i)~modeling and predicting the patterns of change in the environment traversability and (ii)~planning paths that exploit the predictions to reduce the risk of encountering blocked passages.
While existing approaches propose to make decisions according to periodic patterns of change in the environment~\cite{fentanes2015icra}, we focus on modeling spatial patterns.

The main contribution of this paper is a novel system for robot navigation over extended periods of time in indoor environments where changes in traversability follow repeating spatial patterns.
We model the environment by using a topological map like the one illustrated in~\figref{fig:motivation}, where the nodes are the locations of interest and the edges are the traversable passages between these locations.
We incrementally model how the traversability of the edges changes from robot's observations during traversal.
We use a probabilistic graphical model to represent this knowledge and to make predictions of the environment traversability.
We exploit the predictions to plan navigation strategies that account for the risk to encounter obstacles and, at the same time, for the information gain of making observations to improve the model.




As a result of that, our approach is able to
(i) learn incrementally a model of the patterns of change in the environment traversability from robot's observations;
(ii) make predictions about the traversability at unobserved locations;
(iii) plan paths that exploit the predictions to make informed decisions for navigation.
Over time, this leads the robot to encounter a reduced number of blocked passages and, thus, to navigate along paths that are on average shorter than following greedy-reactive strategies. 
%


\section{Related Work}
\label{sec:related}




In the literature, several approaches have been proposed to model changing environments for robot navigation.
For example, Stachniss and Burgard~\cite{stachniss2005aaai} map the typical configurations of low-dynamic areas of the environment for improving localization.
Conditional Transition Maps~\cite{kucner2013iros} offer a grid-based representation for learning the motion patterns of objects in the environment.
Dynamic Gaussian Process Occupancy Maps~\cite{ocallaghan2016iser} map long-term dynamics with a spatially-continuous representation that provides occupancy estimates.
Fremen~\cite{krajnik2017tro} enhances a topological map with a spectral model that allows for predicting the traversability of the edges as a function of the time of day.
Fentanes \etal~\cite{fentanes2015icra} use this model for planning paths that take into account the temporal periodicity of changes in the environment.
In contrast to that, we use a probabilistic approach to learn a time-agnostic model of the traversability changes on the edges of a topological representation of the environment.

Representing the full joint probability of the edge traversability is intractable even for relatively small environments.
In the context of SLAM, FAB-MAP~\cite{cummins2008ijrr} uses the Chow-Liu approximation for modeling the joint distribution of a set of visual features.
For traffic prediction, Furtlehner \etal~\cite{furtlehner2007tits} uses a factor graph representation to predict the road traffic from a probe vehicle.
We also investigate spatial patterns and consider an approximation based on factor graphs~\cite{pearl2014book} which is able to capture the correlation between traversability changes in the environment and exploits this correlation to make probabilistic predictions.

Planning paths in the belief space defined by the predictions can be formulated as a Partially Observable Markov Decision Process~(POMDP). 
However, POMPDs are in practice intractable for real-world environments~\cite{papadimitriou1987mor}.
The Reactive Planning Problem~\cite{macdonald2019ijrr} consider a set of possible configurations and plans policies that guarantee the robot to reach the goal.
Murphy \etal~\cite{murphy2013tra} samples the edge costs from a probabilistic costmap, generates a list of paths using A$^{*}$, and selects the most frequent path.
RAG search~\cite{chung2018ijrr} plans risk-aware paths in a graph where the edge costs are unknown by trading off exploration and exploitation. 
In the Canadian Traveler Problem~(CTP)~\cite{papadimitriou1991tcs}, an agent aims at traveling along the shortest path in a road network where some roads, unknown to the agent, are blocked.
Lim \etal~\cite{lim2017uai} introduce a variant of the CTP in which the roads' traversability may be correlated.
The CTP is a PSPACE-complete problem~\cite{fried2013tcs}, however different approximations have been proposed.
Nikolova and Karger~\cite{nikolova2008aaai} approximate the CTP by considering graphs that consist only of disjoint paths.
Whereas, CTP-UCT~\cite{eyerich2010aaai} is a Monte-Carlo search algorithm that computes policies by taking the uncertainty of the predictions into account.
We extend this approach for planning in our problem paths that lead the robot to collect informative observations for improving the model of the environment traversability and its predictions. 

Krause \etal~\cite{krause2006ipsn} introduce an approach for planning informative paths by selecting the locations that maximize the mutual information~\cite{mackay2004informationbook}.
We use the mutual information for planning paths that trade off the exploration of informative locations and
the exploitation of the traversability predictions to navigate along short paths.

\section{Problem Definition and Assumptions}
\label{sec:problem}

We consider a robot that navigates in indoor environments where the traversability changes according to repeating patterns.
Initially, these patterns are completely unknown to the robot.
It can however observe which passages are frequently blocked at the same time while navigating.

Our robot navigation system relies on a topological map~$\graph\,=\,(\Vertices,\,\Edges\,)$ of the environment, where~$\Edges$ are the edges representing possible paths and $\Vertices$ are the set of nodes representing their intersections. 
We refer to each navigation task performed by the robot as~\emph{run}.
At every run, the traversability of the environment changes. 
We represent the traversability of an edge~$\edge_i$ at a run~$t$ as the binary random variable~$\edge^t_i$ that is~0 if the edge is be blocked or~1 if the edge is free.
We refer to the state of all the edges of the topology during run~$t$ as the \emph{environment configuration}~$\Edges^t\,=\,\{\edge^t_1,\,\dots,\,\edge^t_{|\Edges|}\}$.

In this work, we make the following key assumption:
\begin{enumerate}
\item when a new run starts, the robot has no knowledge about the current environment configuration except from its previous observations;
\item \emph{during} each run, the environment configuration does not change. This means that we account only for the low frequency dynamic changes in the environment;
\item the environment configuration is independent on the temporal order of the runs, i.e. the configuration at run~$t$ has the same degree of dependence to the configuration at run~$t+1$ than the one at~$t+k$;
\item when reaching one of the nodes~$\Vertex\,\in\,\Vertices$, the robot observes all the adjacent edges to~$\Vertex$.
\end{enumerate}




\section{Estimating Patterns in Traversability}
\label{sec:fg}

To navigate in changing environments, we aim at learning a model of the patterns of change to make predictions about the traversability at unknown locations.
We use the robot's observations during traversal to incrementally learn a probabilistic model that captures the correlation among the edge traversabilities and that exploits these correlation to make predictions during navigation.


\subsection{Modeling Environment Configurations}
\label{sec:joint}

\begin{figure*}[t]
\centering
  \vspace{0.2em}
  \subfloat[Example topology.]{
      \includegraphics[height=3.5cm]{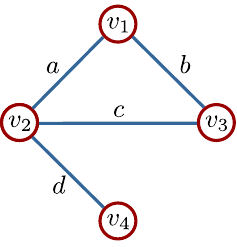}\label{fig:example:topo}}\hspace{2px}
  \subfloat{
      \includegraphics[height=3.7cm]{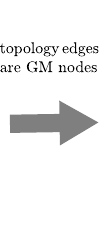}}\hspace{5px}
  \setcounter{subfigure}{1}%
  \subfloat[Indep.\,variables approx.]{
      \includegraphics[height=3.5cm]{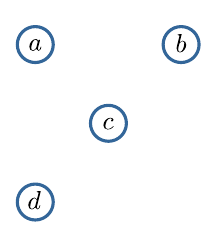}\label{fig:example:naive}}~\hspace{2px}
  \subfloat[Chow-Liu approx.]{
      \includegraphics[height=3.5cm]{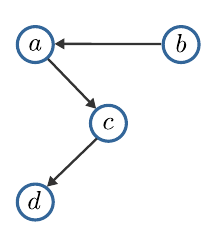}\label{fig:example:chowliu}}~\hspace{2px}
  \subfloat[Our factor graph approx.]{
      \includegraphics[height=3.5cm]{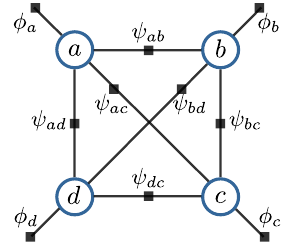}\label{fig:example:fg}}
 \caption{Graphical model~(GM) representations~(b-d) of the joint probability over the edges in the example topology \protect\subref{fig:example:topo}.}
 \label{fig:example:approx}
\end{figure*}

During navigation at run~$t$, the environment presents a configuration~$\Edges^t$ of which the robot only observes a subset of edges~$\ObservedEdges^t\,\subseteq\,\Edges^t$.
To make informed decisions for navigation, we aim at predicting the current environment configuration~$\Edges^t$ and, in particular, the traversability of the unobserved edges~$\UnobservedEdges^t$, with~$\Edges^t\,=\,\ObservedEdges^t\,\cup\,\UnobservedEdges^t$.
Therefore, we look for the probability of the unobserved edges~$\UnobservedEdges^t$ to be traversable 
conditioned on the partial observation of the environment configuration~$\ObservedEdges^t$:
\begin{eqnarray}
  \prob(\UnobservedEdges^t\,\mid\,\ObservedEdges^t)=&\dfrac{\prob(\UnobservedEdges^t,\,\ObservedEdges^t)}{\prob(\ObservedEdges^t)},&{\text{def. cond. prob.}}\label{eq:pred}\\
    =&\eta\;\prob(\Edges^t), &{\Edges^t=\ObservedEdges^t\cup\UnobservedEdges^t}
  \label{eq:joint}
\end{eqnarray}
where~$\eta$ is a normalizer given the current observations~$\ObservedEdges^t$ and~$\prob(\Edges^t)$ is the joint probability distribution over the traversability of the edges in the environment.

The distribution~$\prob(\Edges^t)$ defines a probability function over the space of possible configurations and captures the correlation among the traversability of edges.
This is in general a distribution without a special structure and, thus, the space required to represent it is exponential in the number of edges.
Therefore, representing the joint distribution over the edges becomes quickly intractable.

Different approaches exist to approximate joint probability distributions.
The simplest one is to consider each edge to be independent from all others, as illustrated in~\figref{fig:example:naive} for the example topology in~\figref{fig:example:topo}.
This representation is efficient to store and compute, but it is not able to capture the correlation among the edges.
A more advanced approach is the Chow-Liu approximation~\cite{chow1968tit} that represents the joint probability using a tree-structured Bayesian Network as the one depicted in~\figref{fig:example:chowliu}.
Chow-Liu representation requires quadratic space and is able to capture some correlation among edges.
However, it requires training data and typically does not deal with incremental and partial data which characterize our problem.
To deal with our requirements, we propose to approximate the joint probability distribution over the edge traversability by using a flexible but bounded factor graph representation.




\subsection{Our Factor Graph Model}
\label{sec:dyn:fg}

A factor graph is a probabilistic graphical model that allows for representing a general factorization of a function.
It is structured as an undirected graph with two kinds of nodes: the variable nodes that correspond to the random variables and the factor nodes that represent local functions of the adjacent variable nodes.

We use a factor graph representation in which the variable nodes are the edges~$\Edges$ of the topology.
We model the correlation among the traversability of the edges by defining 
one unary factor node~$\phi$ for each edge and of one binary factor node~$\psi$ for each pair of topology edges, as in the factor graph illustrated in~\figref{fig:example:fg}.
%
Considering this representation, we assume that we can approximate the joint distribution over the edges as:
\begin{eqnarray}
  \prob(\Edges^t)&\approx&p_{\phi\psi}(\Edges^t)~=~\prod_{i}\phi_i\,\prod_{j}\psi_{ij}.
  \label{eq:fg:factors}
\end{eqnarray}

This factor graph representation allows for approximating the probability over the environment configuration
by capturing some of the correlation between the edge traversability while storing only~$|\Edges|(|\Edges|-1)/2+|\Edges|$ low-dimensional factors.
Therefore, our representation requires only quadratic space in the number of edges rather than exponential as in the case of the full joint probability distribution.




\subsection{Learning Factors From Observations}
\label{sec:fg:learning}

Given our factor graph representation, we need to provide a definition for the factors~$\phi$ and~$\psi$ such that
the model corresponds to the robot's observations collected in the previous runs, 
and we can efficiently update the model in an incremental manner as the robot acquires new observations.

The belief propagation algorithm~(BP)~\cite{pearl2014book} allows for performing inference on factor graphs.
We use the BP algorithm to make predictions from current data~(see~\secref{sec:fg:predictions}) but also to estimate the model parameters from robot's observations.
Furtlehner \etal~\cite{furtlehner2007tits} introduce an approach to estimate the factor nodes from the marginal probabilities by using the fixed points of BP algorithm.
We use this approach to define the unary factors and binary factors of our factor graph as:
\begin{eqnarray}
   \phi_i &=&  \prob(\edge_i), 
   \label{eq:dyn:unary}\\
   \psi_{ij} &=& \frac{\prob(\edge_i,\,\edge_j)}{\prob(\edge_i)\,\prob(\edge_j)},
   \label{eq:dyn:binary}
\end{eqnarray}
where~$\prob(\edge_i)$ and~$\prob(\edge_i,\edge_j)$ are respectively the unary and binary joint probabilities of the edges to be traversable or blocked. 
Note that~\eqref{eq:dyn:binary} has an analogy with the mutual information between~$\edge_i$ and~$\edge_j$ that is non-zero for~$\prob(\edge_i,\,\edge_j)\neq\prob(\edge_i)\,\prob(\edge_j)$.
This gives an intuition that such definition of the factors allows for modeling the correlation between edges.

This definition of factor nodes allows for computing the approximated joint probability distribution in~\eqref{eq:fg:factors} as:
\begin{eqnarray}
  p_{\phi\psi}(\Edges^t) &=& 
  \prod_{i}^{|\Edges|}\,\prob(\edge_i)\,\prod_{j}^{|\Edges|}\,\frac{\prob(\edge_i, \edge_j)}{\prob(\edge_i)\,\prob(\edge_j)}.
  \label{eq:fg:probs}
\end{eqnarray}

We compute the unary and binary joint probabilities, $\prob(\edge_i)$ and $\prob(\edge_i,\,\edge_j)$, from the robot's observations in the previous runs~$\ObservedEdges^{1:t\text{-}1}$. 
We achieve this by maintaining a counter of the number of observed occurrences of each unary and binary configurations.
To prevent probabilities to take extreme values of~0 or~1 on a single observation, we initialize them with a uniform prior by assigning to each configuration an equal positive number of occurrences.
After each run, we update the counters based on the robot's observations and recompute the probabilities.
Dealing with unary and binary joint probabilities allows us to update only the probabilities corresponding to the observed edges.
Using this procedure, we can incrementally and efficiently compute the model's parameters from robot's observations.
The other key advantage of this procedure is that it also allows for easily incorporating partial observations of the environment.
For instance, in the example illustrated in~\figref{fig:example:topo}, if the robot observes the edges~$a$ and~$b$ but not~$c$ and~$d$, we update~$\prob(a)$, $\prob(b)$ and~$\prob(a,\,b)$ but not~$\prob(a,\,c)$ and~$\prob(b,\,d)$.

\subsection{Predicting Edge Traversability}
\label{sec:fg:predictions}

Our factor graph model maintains a tractable approximation of the joint probability distribution of the edge traversability, but also provides us a tool for predicting the traversability of currently unknown edges.
The BP algorithm implements a message passing procedure in the graph to estimate the MAP environment configuration and the marginal probabilities of each edge to be traversable.

We predict the traversability of the unobserved edges~$\UnobservedEdges^t$ at run~$t$ by fixing the observed edges~$\ObservedEdges^t$ to the observed values in the factor graph and by performing belief propagation. 
This procedure allows for computing a belief about the environment configuration that approximates~$\probcond{\UnobservedEdges^t}{\ObservedEdges^t}$.
We perform inference on factor graphs by using the implementation of approximate loopy belief propagation of~\emph{libDAI}~\cite{mooij2010jmlr}.
\section{Planning Exploiting Predictions}
\label{sec:planning}

We aim at exploiting the predictions of the edge traversability provided by our factor graph model to plan anticipatory behaviors that lead the robot to encountering a reduced number of unforeseen obstacles during navigation in the long run.
To achieve this, we explore the belief space of possible environment configurations and plan paths that trade off travel distance and information gain to improve the edge traversability model.



\subsection{Estimating Travel Distance From Predictions}
\label{sec:uct}


We estimate the travel distance to reach the goal in a partially observed environment by exploring the belief space of the possible environment configurations defined by the predictions. 
We search the belief space by using an approach based on CTP-UCT~\cite{eyerich2010aaai}.
CTP-UCT is a Monte-Carlo search algorithm based on the Upper Confidence tree algorithm~(UCT)~\cite{kocsis2006ecml} that allows for computing approximate solutions for the Canadian Traveler Problem.


Given a prediction of the environment configuration, we approximate the belief space of possible configurations by performing a sequence of~\emph{rollouts}. 
A rollout randomly samples a configuration according to the current belief and simulates robot navigation based on this configuration.
The robot has no initial knowledge about the sampled configuration but it can make observations during traversal.
To explore the belief space, the robot selects locations for navigation that led to the goal through short paths and have been selected less often in the previous rollouts.
To this end, at each step of the rollouts, we consider a \emph{state} composed by the robot's current location, the set of known traversable and blocked edges, and the set of unknown edges.
Let~$\rho = \{s_0,\,s_1,\,\dots,\,s\}$ be the current sequence of states at the $k$-th rollout, we select the next state~$s'$ that maximizes the UCT formula:
\begin{eqnarray}
  s' = \argmax_{s'}B\sqrt{\frac{\log R^{k\text{-}1}(\rho)}{R^{k\text{-}1}(\rho')}} - dist(s,s') - C^{k\text{-}1}(\rho'), 
\end{eqnarray}
where~$\rho'=\{\rho,s'\}$ is the new sequence of states, 
$B>0$~is a parameter that biases the exploration in the belief space,
$dist(s,\,s')$~is the travel distance to move from~$s$ to~$s'$, 
$R^{k\text{-}1}(\rho)$~is the number of previous rollouts that start with~$\rho$, and
$C^{k\text{-}1}(\rho)$~is the average travel distance to the goal in the previous rollouts that start with~$\rho$.

After performing a number of rollouts, CTP-UCT selects the path~$\mathcal{P}$ that minimizes:
\begin{eqnarray}
  cost_{\,\mathrm{CTP-UCT}}\,(\mathcal{P})&=& {length}(\mathcal{P}),
  \label{eq:dyn:cost}
\end{eqnarray}
where~$length(\mathcal{P})$ is the average travel distance to the goal during the rollouts.
We extend this cost function to plan paths that in the initial runs lead the robot to collect information about the traversability in the environment for improving the model and the predictions in the subsequent runs.





%

\subsection{Collecting Informative Observations}




A common approach to collect information about the environment is to make observations at locations that maximize the mutual information about the non-observed regions.
The mutual information, also called \emph{information gain}, between two discrete random variables~$\var{A}$ and~$\var{B}$ is defined as:
\begin{eqnarray}
  \mutualinfo(\var{A},\,\var{B})  &=& \entropy(\var{A}) - \entropy(\var{A}\,\mid\,\var{B}), \\
            &=& \sum_{a\in \var{A},\,b\in \var{B}}\,\prob(a,\,b)\,\log\frac{\prob(a,\,b)}{\prob(a)\,\prob(b)}\label{eq:dyn:mi},
\end{eqnarray}
where~$\entropy(\cdot)$ and~$\entropy(\cdot\mid\cdot)$ are respectively the entropy and the conditional entropy.

Given the current set of unobserved edges~$\UnobservedEdges^t$, we bias the robot's behavior to collect informative observations by selecting paths that maximize the information gain:
\begin{eqnarray}
  \mutualinfo(\varset{P},\,\UnobservedEdges^t) &=& \entropy(\varset{P}) - \entropy(\varset{P}\,\mid\,\UnobservedEdges^t),
\end{eqnarray}
where~$\varset{P}$ are the edges along the path~$\mathcal{P}$.

Computing the entropy over~$\varset{P}$ requires the knowledge of their joint probability but our factor graph model does not provide direct access to it.
Therefore, we approximate~$\mutualinfo(\varset{P},\UnobservedEdges^t)$ with the sum of the pairwise mutual information between the edges along~$\mathcal{P}$ and the unobserved edges~$\UnobservedEdges^t$:
\begin{eqnarray}
  \quad\mutualinfo(\varset{P},\,\UnobservedEdges^t)~\approx~\hat{\mutualinfo}(\varset{P},\,\UnobservedEdges^t)~=\,\sum_{U^t\in\UnobservedEdges^t}\,\max_{P\in\varset{P}}\,\mutualinfo(P,\,U^t),
  \label{eq:dyn:MI}
\end{eqnarray}
where~$\mutualinfo(P,\,U^t)$ is computed using~\eqref{eq:dyn:mi} and involves only unary and binary joint probabilities that are directly available from our factor graph model.
We make sure that the mutual information for the same edge is not counted multiple times by considering the maximum mutual information for each unobserved edge~$U^t$.


\subsection{Exploration Vs. Exploitation}
We aim at exploring the environment configurations while minimizing the travel distance.
To this end, we select paths according to a cost function that trades off the the path length and the information gain along it:
\begin{eqnarray}
  cost(\mathcal{P})~=~length(\mathcal{P}) - \gamma^{\mathrm{\#runs}}\,\left[\zeta\;\hat{\mutualinfo}(\varset{P},\,\UnobservedEdges^t) \right],
  \label{eq:dyn:cost}
\end{eqnarray}
where~$\varset{P}$ are the edges along~$\mathcal{P}$, $length(\mathcal{P})$ is the average travel distance to reach the goal following~$\mathcal{P}$ in the rollouts, $\gamma\in [0,1]$ is a parameter that controls the exploration term, and~$\zeta$ is a constant that normalizes the information gain with respect to the travel distance.


The exploratory behavior of the robot is determined by the parameter~$\gamma$ that decays exponentially with the number of runs performed by the robot.
Initially, when few observations are available, $\gamma$~leads the robot to favor exploratory behaviors for improving the model of the traversability of the edges.
As the robot performs a number of runs and acquires several observations of the environment, the model and its ability to make predictions improve and the exploration becomes less and less prominent.
When the learning process of the model converges, our problem becomes similar to a CTP in which the traversability of the edges is correlated. 
At this point, the exploratory term in~\eqref{eq:dyn:cost} is weighted low and, thus, the cost function becomes similar to the original CTP-UCT.

\section{Experimental evaluation}
\label{sec:dyn:exp}

The main focus of this work is on robot navigation over extended periods of time in environments where the traversability changes by following repeating patterns and on how the knowledge about such patterns can be obtained and exploited.
Our experiments are designed to illustrate that our approach is able to
(i)~model and make predictions about the traversability changes in the environment from the robot's incremental observations;
(ii)~plan paths that exploit the predictions to make informed decisions for navigation;
(iii)~navigate along paths that are on average shorter than following greedy shortest path strategies.


In our experiments, we consider different topologies defined over real-world environments.
There exist many approaches to build topological maps for example Kuipers \etal~\cite{kuipers2004icra}.
On these topologies, we simulate repeating patterns in the traversability changes.
First, we sample~$M$ independent template configurations for each environment.
Then, we generate a set of~$N,\,N \gg M$ configurations by sampling uniformly one of the~$M$ templates and applying random noise on edge traversability.

\subsection{Predicting Environment Configurations}

\begin{figure}[t]
\vspace{0.4em}
\centering
\includegraphics[width=0.86\linewidth]{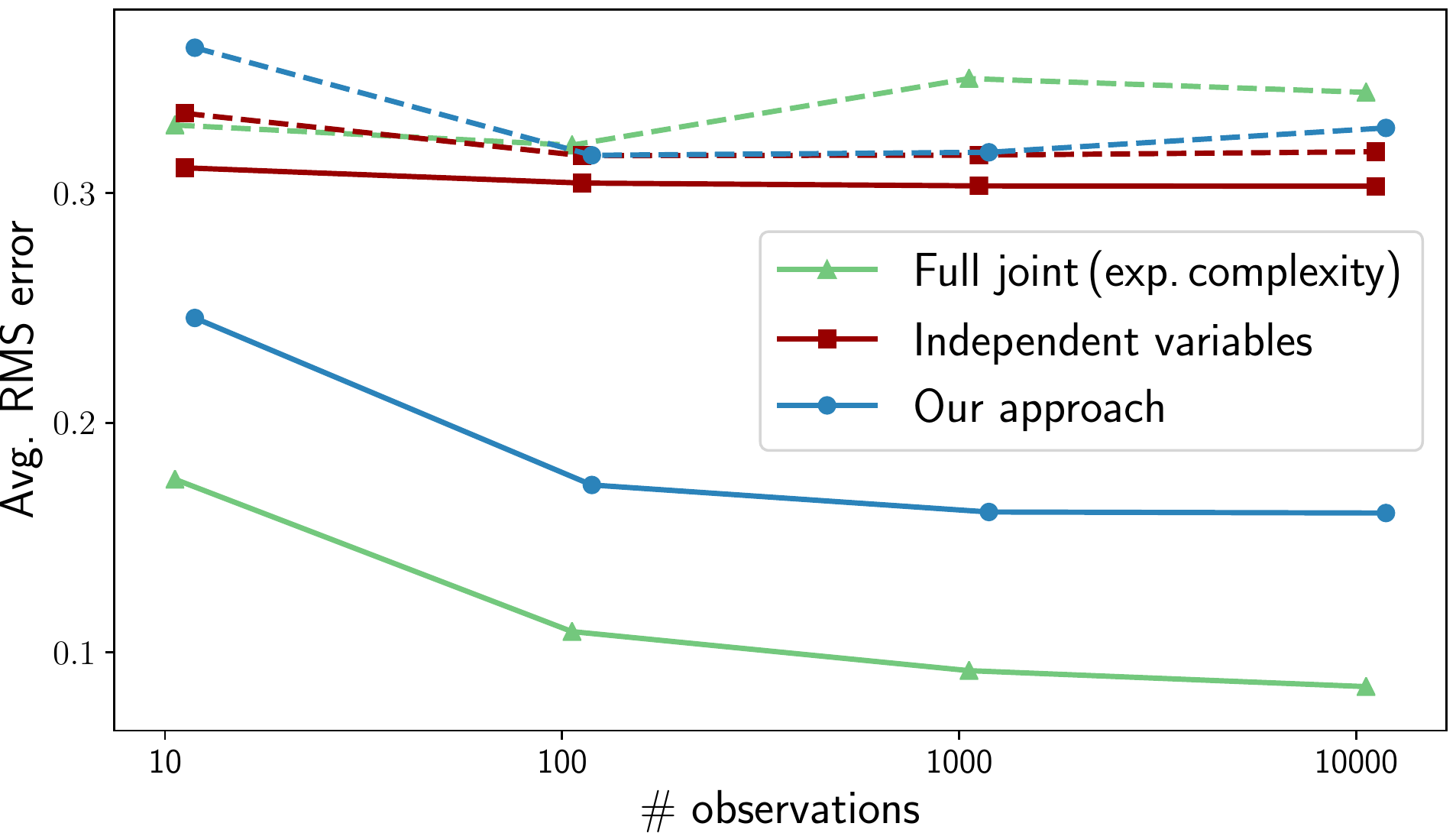}
\includegraphics[height=4.125cm]{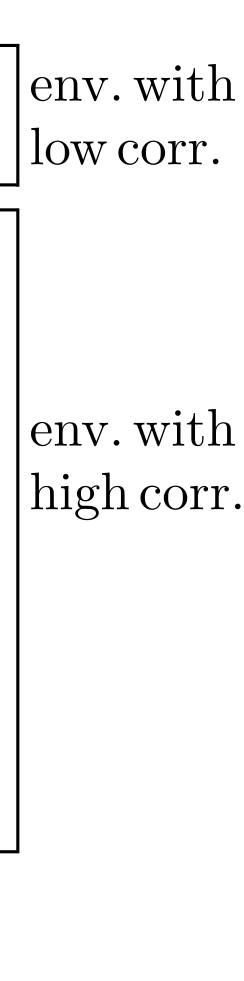}
\caption{Avg. RMS error of the predictions for low and high correlated environment configurations using different models.}
\label{fig:exp:predictions}
\end{figure}

We designed the first experiment to show the capabilities of our approach to model the traversability changes in the environment and to make predictions about the environment configurations.
In this experiment, we consider a relatively small topology composed by~9 nodes and~13 edges and assume that the robot observes at each run the whole environment.
These assumptions allows for computing the ground truth joint distribution over the edge traversability despite its exponential space complexity to use for comparison.
We additionally compare our approach with a model that assumes each edge to be independent from all the others as the one illustrated in~\figref{fig:example:naive}.
To investigate the performance of our approach, we consider two different cases: one in which the environment configurations are highly correlated, and one in which the correlation is low.

We compare the capabilities of each model to predict the edge traversability for~10000 partial configurations after the robot observed~10, 100, 1000, and~10000 configurations.
In~\figref{fig:exp:predictions}, we illustrate the RMS error of the predictions to the true configurations for the three approaches.
In cases in which the configurations are highly correlated~(solid lines), our approach~(blue) is able to provide good predictions already after few observations.
The predictions improve incrementally as the robot makes more observations, similarly as if using the full joint probability distribution~(green).
Instead, assuming the edges to be independent~(red) cannot capture the correlation between edges and leads to worse predictions.
In cases in which the configurations have low correlation~(dotted line), all of the three approaches provide similar predictions. 
Therefore, also in situations with low correlation, our approach does not reveal worse performance than the model assuming independence among edges.

\subsection{Navigation Exploiting Predictions}
\label{sec:exp:planning}

The second experiment is designed to show that our approach is able to exploit the predictions of the environment configurations to plan informed strategies that lead the robot to navigate along shorter paths over time.
In this experiment, we consider four different environments described in~\tabref{tab:environments}.
To evaluate the improvement over time, we repeat a fixed sequence of navigation of~25 tasks and configurations for a total of~500 runs in each environment.
We compare our approach to the theoretical optimal path computed in the ground truth environment configuration (unknown to the robot) and to an optimistic shortest path policy called~\emph{SPO}.
This strategy plan paths using A$^*$ by assuming that every edge of the environment is traversable unless the robot observes the opposite and re-plans.
SPO does not take into account the predictions of the environment configuration and thus, independent of the number of runs, makes the same decisions as if navigating for the very first time.
In our approach, we perform 50~rollouts per decision and set the parameter that regulates the exploratory term of the cost function to~$\gamma = 0.95$.

\begin{figure}[t]
  \vspace{0.4em}
  \centering
  \includegraphics[width=0.86\linewidth]{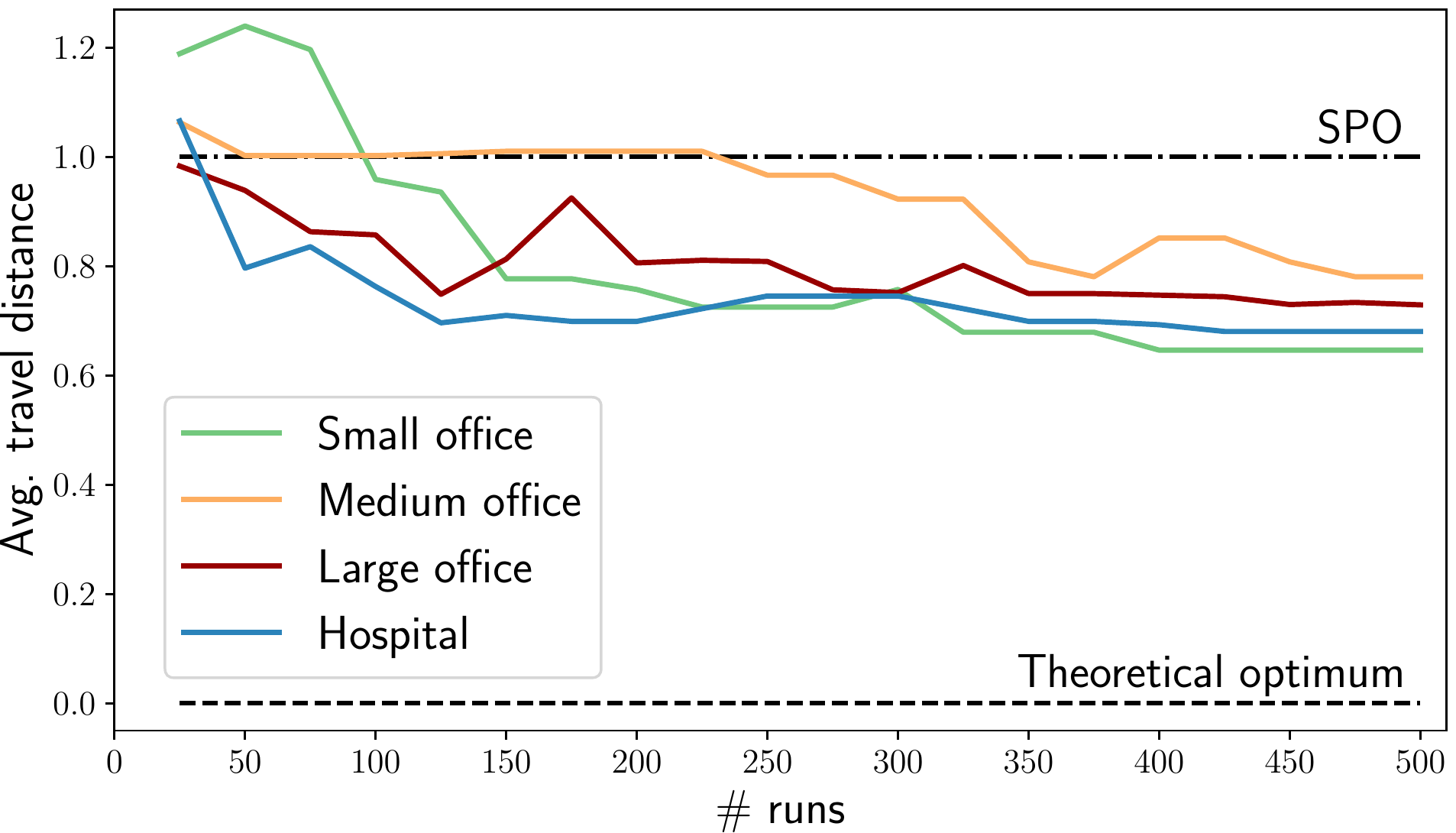}
  \caption{Avg. distance traveled by the robot following our approach over the number of runs in the environments described in~\tabref{tab:environments} normalized between the theoretical optimum~(black dashed line) and the SPO solution~(black dash-dotted line).}
  \label{fig:exp:costnav}
\end{figure}

\begin{table}[t]
  \centering
  \caption{Environments considered in our experimental evaluation.}
  \begin{tabular}{@{}lcccc@{}}
  \toprule
  \multicolumn{1}{l}{Environment} & \multicolumn{1}{l}{Dimensions} & \multicolumn{1}{l}{Nodes} & \multicolumn{1}{l}{Edges} \\ \midrule
  \multicolumn{1}{l|}{Small office} & 25 $\times$ 20 m & 16 & 18  \\
  \multicolumn{1}{l|}{Medium office} & 30 $\times$ 30 m & 20 & 30  \\
  \multicolumn{1}{l|}{Large office} & 50 $\times$ 30 m & 18 & 37  \\
  \multicolumn{1}{l|}{Hospital} & 125 $\times$ 35 m & 40 & 55 \\ \bottomrule
  \end{tabular}
  \label{tab:environments}
\end{table}

The performance of our approach over the number of runs in the four environments are illustrated in~\figref{fig:exp:costnav}.
We evaluate the average difference in the travel distance to the theoretical optimum with ground truth environment knowledge available~($0.0$) normalized with respect to the SPO solution~($1.0$).
Initially, when the robot collected little information about the environment, the predictions of the edge traversability are weak and the robot following our approach performs similarly as following SPO.
After~100 runs, the robot starts discovering patterns in the traversability changes and our approach is able to plan paths leading it to the goal along shorter paths.
Over time, when the robot collects more and more observations about the environment, the learning process of the edge traversability model converges and, after 500~runs, the robot is able to navigate along paths that are on average about~30\% shorter than following an optimistic shortest path strategy.

\subsection{Planning Performance Comparison}

\begin{figure}[t]
  \vspace{0.4em}
  \centering
  \includegraphics[width=0.86\linewidth]{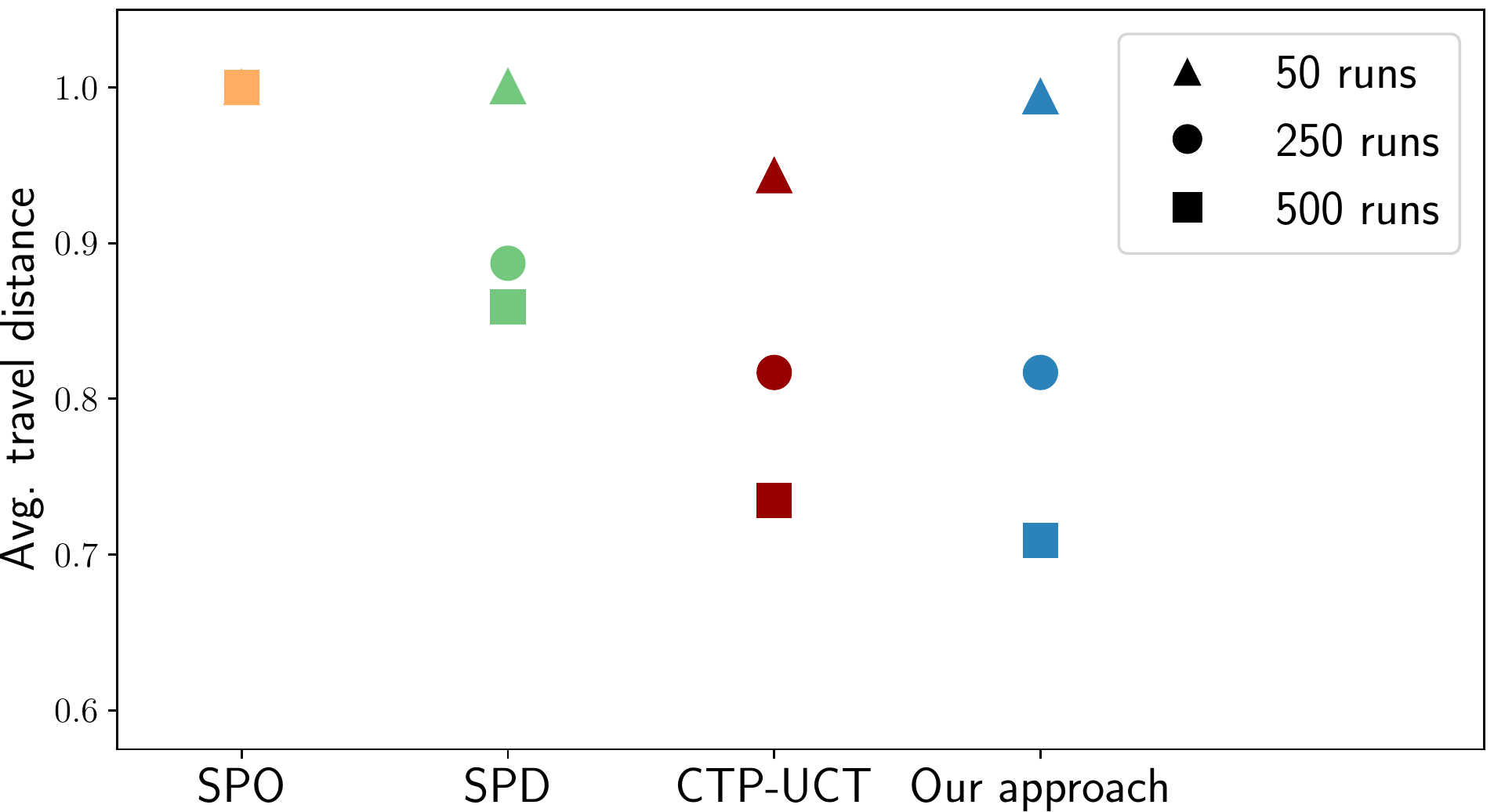}
  \caption{Avg. distance traveled by the robot following different planning approaches. 
  CTP-UCT and SPD use our factor graph model for computing the predictions.}
  \label{fig:exp:predsnav}
\end{figure}

Besides the baselines discussed in the previous section, we compare the performance of our approach to other planners.
For comparison, we consider the original~\emph{CTP-UCT}~\cite{eyerich2010aaai} that searches for the shortest path in the predicted belief as described in~\secref{sec:uct}.
We compare our approach also to a strategy inspired by Lim \etal~\cite{lim2017uai} called~\emph{SPD}.  
SPD makes a most likely assumption on the belief about the edge traversabilities and plans the shortest path using A$^*$ on the determinized environment configuration.
When the robot makes an observation incompatible with the current determinized configuration, SPD computes a new prediction and re-plans.

It is important to note that CTP-UCT and SPD do not provide an approach to model and make predictions of the edge traversability in the environment.
Therefore, we consider the predictions provided by our factor graph model also for these approaches.

The average performance of the approaches after~50, 250 and 500 runs for navigating in the environments introduced in the previous section are illustrated in~\figref{fig:exp:predsnav}.
SPO~(orange) reveals a constant trend over time as it does not take the predictions into account.
Considering the predictions of the edge traversability, SPD~(green) leads the robot to navigate along shorter paths over time.
However, the determinization of the predicted configurations may cause the robot to follow paths that are distant from the optimal ones.
CTP-UCT~(red) consider a weaker approximation of the belief defined by the predictions by performing rollouts.
Thus, it is able to make more informed decisions than SPD that lead the robot along shorter paths.
Our approach~(blue) extends CTP-UCT by considering an exploratory term that allows the robot to collect informative observations that explicitly improve the model and so the predictions of the edge traversabilities.
The exploratory behavior leads initially to slightly longer travel distances than CTP-UCT but, in the long run, it allows the robot to navigate along shorter paths.

\section{Conclusion}
\label{sec:conclusion}

In this paper, we investigate robot navigation over extended periods of time in environments where the traversability changes according to repeating patterns.
We present an approach that learns a probabilistic model of the traversability changes from robot's incremental observations during navigation. 
Our model predicts the traversability at unknown locations exploiting the estimated correlation between the traversability in the environment.
We exploit these predictions to make informed decisions that lead the robot to navigate by encountering a reduced number of unforeseen obstacles.

Although our approach presents a higher complexity in comparison to traditional planning systems,
in environments where the traversability changes following certain patterns, it has the potential to lead robots to automatically navigate along shorter paths over time increasing the efficiency of their operations.








\clearpage
\bibliographystyle{plain}

\bibliography{glorified,new}

\end{document}